\documentclass{article}
\PassOptionsToPackage{table}{xcolor}

\usepackage{arxiv}
\usepackage{CJKutf8}
\usepackage[utf8]{inputenc} 
\usepackage[T1]{fontenc}    
\usepackage{hyperref}       
\usepackage{url}            
\usepackage{booktabs}       
\usepackage{amsfonts}       
\usepackage{amsmath}
\usepackage{nicefrac}       
\usepackage{microtype}      
\usepackage{lipsum}
\usepackage{graphicx}
\usepackage{xcolor}
\usepackage{natbib}
\usepackage{multirow}
\usepackage{array}
\usepackage[table]{xcolor}
\usepackage{pifont}
\usepackage[most,breakable]{tcolorbox} 
\graphicspath{ {./images/} }
\usepackage{color, soul}
\usepackage{subcaption}

\title{MagicGUI-RMS: A Multi-Agent Reward Model System for Self-Evolving GUI Agents via Automated Feedback Reflux}

\author{
	\normalfont
	Zecheng Li\thanks{These authors contributed equally to this work.},\,\,
	Zhihui Cao\footnotemark[1]\,\,\thanks{Corresponding authors},\,\,
	Wenke Huang,\,\,
	Yudong Zhang,\,\,
	Keying Qi,\,\,
	Rui Wang,\,\,
	Zeyu Zheng\\
	Jian Zhao,\,\,
	Hao Zhu,\,\,
	Hengxin Wu,\,\,
	Yuran Wang,\,\,
	Guitao Fan,\,\,
	Guokun Wu,\,\,
	Yicong Liu,\,\,
	Zhilin Gao\\
	Haikun Xu,\,\,
	He Yang,\,\,
	Minqi Xiang,\,\,
	Xingyu Liu\footnotemark[2],\,\,
	Zuojian Wang\footnotemark[2]\\
	\\[0.4em]
	\textit{Honor Device Co., Ltd}
}

\begin{document}
\maketitle

\begin{abstract}
    \renewcommand{\thefootnote}{\fnsymbol{footnote}}
Graphical user interface (GUI) agents are rapidly progressing toward autonomous interaction and reliable task execution across diverse applications. However, two central challenges remain unresolved: automating the evaluation of agent trajectories and generating high-quality training data at scale to enable continual improvement. Existing approaches often depend on manual annotation or static rule-based verification, which restricts scalability and limits adaptability in dynamic environments.
We present MagicGUI-RMS, a multi-agent reward model system that delivers adaptive trajectory evaluation, corrective feedback, and self-evolving learning capabilities. MagicGUI-RMS integrates a Domain-Specific Reward Model (DS-RM) with a General-Purpose Reward Model (GP-RM), enabling fine-grained action assessment and robust generalization across heterogeneous GUI tasks. To support reward learning at scale, we design a structured data construction pipeline that automatically produces balanced and diverse reward datasets, effectively reducing annotation costs while maintaining sample fidelity. During execution, the reward model system identifies erroneous actions, proposes refined alternatives, and continuously enhances agent behavior through an automated data-reflux mechanism.
Extensive experiments demonstrate that MagicGUI-RMS yields substantial gains in task accuracy, behavioral robustness. These results establish MagicGUI-RMS as a principled and effective foundation for building self-improving GUI agents driven by reward-based adaptation.
\renewcommand{\thefootnote}{\arabic{footnote}}
\end{abstract}

\section{Introduction}

As multimodal large language models (MLLMs) advance toward interactive and embodied intelligence, graphical user interface (GUI) agents have become a pivotal paradigm bridging perception, reasoning, and execution within digital environments~\citep{schneider2024foundation,chen2024internvl,Cogvlm}. By perceiving on-screen visual information, interpreting the semantics and functions of interface elements, and generating executable actions, GUI agents enable end-to-end automation across mobile and desktop applications. This integration of vision and language understanding signifies a paradigm shift from static rule-based automation to adaptive, general-purpose digital intelligence. Recent progress in vision–language foundation models (e.g., GPT-4o~\citep{hurst2024gpt}, Qwen2.5-VL~\citep{Qwen2-VL}) has dramatically enhanced agents’ capabilities in interface grounding, multimodal reasoning, and task planning under uncertainty, accelerating their evolution from task-specific tools to autonomous and generalizable operators capable of complex cross-application control~\citep{wang2024gui, Qwen2-VL, nguyen2024gui}.

Driven by recent advances, research on GUI agents has progressed along three main trajectories~\citep{li2025survey}. Prompt-driven systems such as AppAgent~\citep{zhang2025appagent} and Mobile-Agent~\citep{wang2024mobile} leverage large commercial MLLMs for visual–linguistic reasoning via prompt engineering, achieving strong zero-shot performance but exhibiting limited flexibility due to handcrafted prompts and closed APIs. Supervised fine-tuning (SFT) approaches including CogAgent~\citep{hong2024cogagent}, OS-Atlas~\citep{wu2024atlas}, UI-TARS~\citep{qin2025ui}, and UGround~\citep{qian2025uground} 
train unified perception, grounding, and execution models on large-scale GUI datasets~\citep{li2025screenspot, rawles2024androidworld, hsiao2025screenqa, lu2025guiodyssey}, substantially enhancing screen understanding and sequential reasoning; however, their reliance on static datasets hinders adaptation to dynamic layouts and unseen interfaces. Reinforcement learning and reward-based methods have recently emerged as a promising direction for feedback-driven improvement. Inspired by OpenAI o1~\citep{jaech2024openai} and DeepSeek-R1~\citep{guo2025deepseek}, systems such as DigiRL~\citep{bai2024digirl}, DistRL~\citep{wang2024distrl}, VLM-R1~\citep{shen2025vlm}, and Visual-RFT~\citep{liu2025visual} incorporate structured reward signals to strengthen decision robustness and grounding accuracy. Extending this paradigm, UI-R1~\citep{lu2025ui} applies R1-style RL to GUI action prediction, GUI-R1~\citep{luo2025gui} generalizes reward-guided reasoning across heterogeneous platforms, UI-Genie~\citep{xiao2025ui} introduces a reward-model-centric self-improvement loop.

Despite recent progress, several core limitations remain unresolved~\citep{li2025survey, wang2024gui, nguyen2024gui}. Existing reward mechanisms predominantly offer coarse binary signals (e.g., correct vs.\ incorrect) and lack fine-grained diagnostics regarding failure sources, deviation patterns, or interface-grounding errors, limiting their utility for actionable correction. Trajectory evaluation and action policy generation also remain structurally decoupled, with no unified mechanism that aligns error detection with the refinement of subsequent decisions. Moreover, current trajectory construction and reward-modeling pipelines rely heavily on manually crafted heuristics or single-model judgments, constraining scalability and continual adaptation in real-world environments. Taken together, these limitations hinder reliable self-evolution and underscore the need for a unified reward modeling framework that enables interpretable evaluation, trajectory-level correction, and continual learning.

To address these limitations, we present MagicGUI-RMS, a multi-agent reward modeling framework that unifies adaptive evaluation, corrective feedback, and continual self-evolution for improving GUI agents. MagicGUI-RMS incorporates a Domain-Specific Reward Model (DS-RM) and a General-Purpose Reward Model (GP-RM), coupling fine-grained operational knowledge with broad semantic reasoning across diverse interfaces. This collaborative architecture enables accurate, interpretable, and transferable trajectory assessment in heterogeneous GUI environments.
Furthermore, we develop a structured reward data construction pipeline that produces balanced training samples through trajectory perturbation, rule-based validation, and difficulty-aware sampling, substantially reducing annotation overhead while preserving diversity. During execution, a reward-driven trajectory correction procedure transforms the reward model from a passive evaluator into an active controller by delivering real-time corrective signals. Finally, an automated data-reflux mechanism continually verifies, filters, and reuses high-quality trajectories, enabling sustained performance gains and long-horizon self-evolution through iterative retraining.

In summary, this paper makes the following key contributions:

\begin{itemize}
	\item \textbf{Multi-Agent Reward Model System.} We introduce MagicGUI-RMS, a multi-agent reward modeling framework that jointly leverages a domain-specific reward model and a general-purpose reward model. By coupling task-specialized operational priors with broad semantic reasoning capabilities, the system achieves robust, accurate, and generalizable trajectory evaluation across heterogeneous tasks and application domains.
	\item \textbf{Structured Synthetic Reward-Data Pipeline.} We develop a scalable reward data synthesis pipeline that automatically generates balanced and diverse supervision signals, reducing reliance on manual annotation and providing high-quality data crucial for training reliable reward models.
	\item \textbf{Reward-Guided Trajectory Correction.} We design a step-level corrective feedback loop in which reward signals actively steer trajectory updates, yielding more reliable execution and more stable optimization dynamics.
	\item \textbf{Continual Self-Evolution Mechanism.} We introduce an automated data reflux mechanism that continuously incorporates verified trajectories into the training process, enabling iterative model refinement and sustained performance improvement over time.
\end{itemize}

Together, these innovations position MagicGUI-RMS as a unified, feedback-driven learning paradigm in which adaptive evaluation, trajectory correction, and continual self-evolution jointly drive progressive capability formation, fostering autonomous, generalizable, and reliable GUI-agent capabilities in real-world environments.
\section{Related Work}

\subsection{Prompt-Driven GUI Agents}
The rise of high-capacity Multimodal Large Language Models (MLLMs) has enabled early progress in GUI automation by allowing agents to interpret on-screen visual information and produce executable actions through prompt-based control. Early frameworks such as AppAgent~\citep{zhang2025appagent} and Mobile-Agent~\citep{wang2024mobile} utilized commercial proprietary models (e.g., GPT-4o~\citep{hurst2024gpt}, Gemini~\citep{team2023gemini}) and relied heavily on manually crafted prompts to support device control, task planning, and interface understanding across mobile and desktop environments. While these prompt-driven systems demonstrated strong zero-shot performance, they remained fundamentally constrained by the brittleness of handcrafted prompts and lacked the adaptive reasoning capabilities required to generalize to previously unseen or highly specialized GUI tasks.

\subsection{Supervised Fine-Tuning-Based GUI Agents}
To mitigate the reliance on prompt engineering, subsequent research introduced Supervised Fine-Tuning (SFT) on curated multimodal datasets to improve grounding and action execution. Several end-to-end frameworks have been proposed to advance domain-specific GUI agent capabilities. SeeClick~\citep{cheng2024seeclick} and UGround~\citep{qian2025uground} enhance UI element recognition through dense captioning and hierarchical grounding, while Aria-UI~\citep{yang2025aria} unifies visual perception and action generation within a single MLLM pipeline. In parallel, OS-Atlas~\citep{wu2024atlas} establishes a cross-platform GUI grounding benchmark, and GUI-Odyssey~\citep{lu2025guiodyssey} provides fine-grained supervision over click targets, textual attributes, and action semantics.
UI-TARS~\citep{qin2025ui} further advances this line of work by introducing a native end-to-end agent architecture that integrates perception, reasoning, and memory through system-2 reflective loops trained on large-scale screenshot corpora. More recent large-model frameworks such as MagicGUI~\citep{tang2025magicgui}, CogAgent~\citep{hong2024cogagent} strengthen multimodal grounding via hybrid pretraining on synthetic GUI datasets and spatially structured visual encoders.
Despite these advancements, SFT-based approaches remain limited by their dependence on static supervision, restricting their capacity for continual self-improvement and hindering adaptation to dynamic, real-world GUI environments.

\subsection{Rule-Driven Reinforcement Fine-Tuning-Based GUI Agents}
Rule-driven reinforcement fine-tuning provides a scalable alternative to human-annotated supervision by leveraging symbolically verifiable reward functions. Early work demonstrated that rule-based RL can substantially enhance generalization across reasoning and multimodal tasks by enforcing deterministic constraints such as action-type consistency, coordinate correctness, and IoU-based semantic alignment~\citep{jaech2024openai, guo2025deepseek, shen2025vlm, liu2025visual}.

For mobile GUI tasks, this paradigm has been widely adopted to mitigate the high cost of collecting task demonstrations. Online RL methods such as DigiRL and DistRL~\citep{bai2024digirl, wang2024distrl} gather interaction trajectories in simulated environments and rely on an auxiliary Vision–Language Model (VLM) to determine task completion, forming a rule-driven reward signal that guides policy updates during exploration. In parallel, the emergence of large-scale static GUI datasets~\citep{cheng2024seeclick, li2025screenspot, chen2025guicourse, li2024effects, lu2025guiodyssey} has enabled offline rule-based training at scale, reducing inference overhead and improving sample efficiency. Systems trained on large-scale GUI datasets, such as ReachAgent~\citep{wu2025reachagent}, UI-R1~\citep{lu2025ui}, and GUI-G1~\citep{luo2025gui}, apply deterministic rules to validate step-level correctness and supervise offline policy refinement. More advanced variants, including GUI-R1~\citep{luo2025gui} and InfiGUI-R1~\citep{liu2025infigui}, further leverage these datasets to incorporate rule-guided GRPO/RLOO, enabling joint improvements in grounding fidelity and high-level task execution.

Despite these advantages, rule-driven approaches remain constrained by the expressiveness of handcrafted rules. They primarily capture low-level syntactic and spatial correctness, making it difficult to model nuanced interface semantics or provide fine-grained feedback in complex, multi-step GUI tasks. As a result, rule-based RL often struggles with scalability and generalization in real-world, high-diversity application environments.

\subsection{Reward-Model-Driven Reinforcement Fine-Tuning-Based GUI Agents}
Recent research has increasingly focused on reinforcement fine-tuning driven by reward models, where learned evaluators replace manually designed rules and provide fine-grained, scalable feedback for step-level actions or full trajectories. UI-Genie~\citep{xiao2025ui} introduced UI-Genie-RM, a multimodal reward model capable of assessing action correctness and task outcomes with dual-granularity signals. 
Meanwhile, WEBRL~\citep{qi2024webrl} proposed an online curriculum reinforcement learning framework guided by Outcome-Supervised Reward Models. Through the automatic generation of new tasks from failed rollouts and policy updates regularized by KL constraints, WEBRL enabled open-source models such as GLM-4~\cite{glm2024chatglm} and Llama-3.1~\citep{vavekanand2024llama} to surpass GPT-4o~\citep{hurst2024gpt} on the WebArena-Lite benchmark~\citep{zhou2023webarena}. 

These efforts collectively indicate a transition from static rule-based reinforcement fine-tuning toward reward-model-driven continual learning. By enabling evaluators to diagnose error patterns, provide nuanced corrective signals, and evolve jointly with the policy, this line of research offers a scalable pathway toward autonomous, generalizable, and semantically aligned GUI agents capable of adapting to diverse and dynamic real-world environments.

\section{Methodology}
This section presents MagicGUI-RMS, a multi-agent reward model system designed to enhance GUI agents through adaptive evaluation, corrective feedback, and automated data reflux.
Sec.~\ref{sec:method-rms} introduces the overall system architecture, including two complementary components of the system, the Domain-Specific Reward Model (DS-RM) and the General-Purpose Reward Model (GP-RM).
Sec.~\ref{sec:method-datacon} describes the data construction pipeline for reward modeling, which constructs difficulty-aware positive and negative samples to support fine-grained reward supervision.
Finally, Sec.~\ref{sec:method-reflux} explains the reward-guided data reflux and co-evolution mechanism that iteratively improves both reward models and agent policies.

\begin{figure}[tbp]
	\centering
	\includegraphics[width=0.95\textwidth]{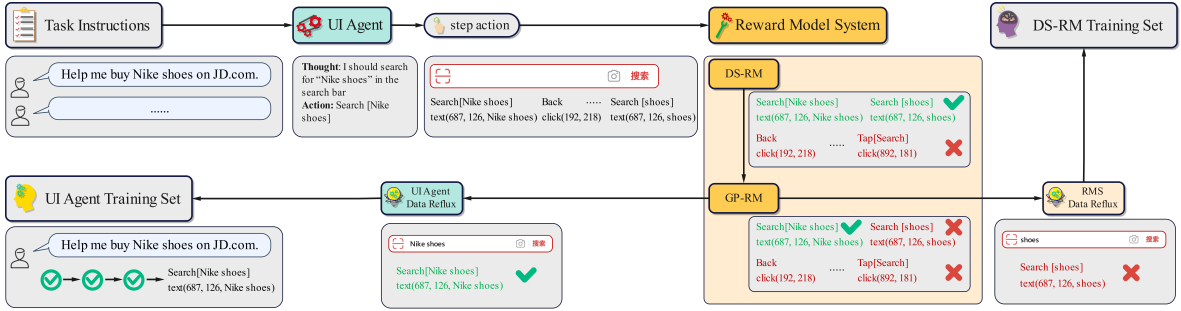}
	\caption{MagicGUI-RMS architecture. The system operates in a three-stage pipeline: (1) the UI Agent proposes a step-level action conditioned on the task instruction and screen state; (2) the action undergoes hierarchical assessment by DS-RM and GP-RM within the Reward Model System; and (3) two complementary data-reflux loops iteratively improve the reward models and the UI Agent.}
	\label{fig:magicgui-rms framework}
\end{figure}

\subsection{MagicGUI-RMS: A Multi-Agent Reward Model System}
\label{sec:method-rms}

MagicGUI-RMS is a hierarchical reward modeling framework that evaluates agent actions through domain-grounded reasoning and global semantic verification. As illustrated in Fig.~\ref{fig:magicgui-rms framework}, the framework operates in three sequential stages:  
(1) action proposal,  
(2) hierarchical reward evaluation, and  
(3) dual-loop data reflux that supports continual self-improvement.

Given the task instruction $x$, the screen state $s$, and the historical trajectory $h_{1:t-1}$, the UI Agent produces an action proposal $a_{\text{pred}}$
\[
a_{\text{pred}} = \pi_{\text{Agent}}(x, s, h_{1:t-1}).
\]
This predicted action is then passed into the hierarchical reward evaluation pipeline.

The evaluation begins with the Domain-Specific Reward Model (DS-RM), which checks whether $a_{\text{pred}}$ satisfies deterministic UI interaction constraints and generates a corrected action $a_{\text{corr}}$ when necessary. The resulting candidate actions $\{a_{\text{pred}}, a_{\text{corr}}\}$, together with DS-RM’s binary decision (correct vs. incorrect) and decision rationale, are forwarded to the General-Purpose Reward Model (GP-RM). GP-RM verifies DS-RM’s decision under broader semantic and contextual considerations, ensuring coherence with global task intent and long-horizon dependencies.

MagicGUI-RMS incorporates two complementary data reflux loops. RMS Data Reflux collects disagreement cases between DS-RM and GP-RM to refine reward model decision boundaries. UI Agent Data Reflux stores the final GP-endorsed action in the UI Agent Training Set as high-quality supervision. Through repeated cycles of action prediction, hierarchical reward evaluation, and data reflux, both reward models and the UI Agent progressively co-evolve toward higher accuracy, robustness, and semantic coherence.

\paragraph{Domain-Specific Reward Model}

The Domain-Specific Reward Model (DS-RM) evaluates whether the agent's predicted action conforms to deterministic UI interaction rules. Given the input
\[
z_{\text{DS}} = (x, s, a_{\text{pred}}, h_{1:t-1}),
\]
DS-RM produces
\[
(y_{\text{DS}}, r_{\text{DS}}, a_{\text{corr}}, r_{\text{corr}}) = f_{\text{DS}}(z_{\text{DS}}),
\]
where $y_{\text{DS}}$ denotes a binary correctness label under domain rules; $r_{\text{DS}}$ provides the rationale for the decision; $a_{\text{corr}}$ is the corrected action produced when $y_{\text{DS}} = 0$; and $r_{\text{corr}}$ offers a brief explanation of the correction.

DS-RM is trained on a domain-aligned reward dataset with correctness annotations and structured error patterns, enabling precise modeling of UI layouts, functional regions, and application-specific constraints. Disagreement-based data reflux further refines DS-RM during system co-evolution.

\paragraph{General-Purpose Reward Model}

The General-Purpose Reward Model (GP-RM) provides semantic and contextual validation beyond deterministic domain logic. It receives the augmented input
\[
z_{\text{GP}} = (x, s, a_{\text{pred}}, h_{1:t-1}, y_{\text{DS}}, r_{\text{DS}}, a_{\text{corr}}, r_{\text{corr}}),
\]
and outputs
\[
(y_{\text{GP}}, e_{\text{GP}}, s_{\text{GP}}) = f_{\text{GP}}(z_{\text{GP}}),
\]
where $y_{\text{GP}}$ denotes the semantic verification of DS-RM’s decision; $e_{\text{GP}}$ provides a binary judgment indicating whether the task has been completed; and $s_{\text{GP}}$ represents an action-level preference used to guide the agent.

GP-RM serves as a global arbiter that evaluates DS-RM’s outputs under broader task semantics, unseen layouts, and long-horizon dependencies.

\begin{figure}[tbp]
	\centering
	\includegraphics[width=0.95\textwidth]{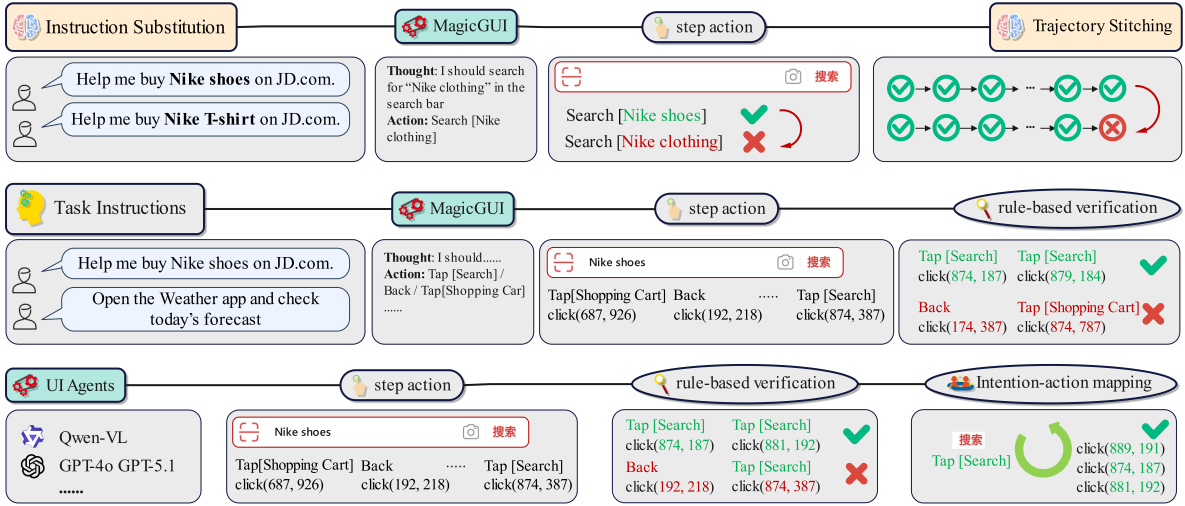}
	\caption{Overview of the reward data construction pipeline. (1) Structured perturbations, including instruction substitution and trajectory stitching, introduce controllable inconsistencies that yield easy negatives. (2) Rule-based verification evaluates MagicGUI actions under standard instructions to produce positive and hard-negative samples. (3) Intention-centric grounding correction refines actions from open-source UI Agents, generating positive samples from intention-aligned behaviors and moderate negatives.}
	\label{fig:data-pipeline}
\end{figure}

\subsection{Data Construction for Reward Modeling}
\label{sec:method-datacon}

To enable low-cost, scalable, and structured reward data synthesis for training the Domain-Specific Reward Model (DS-RM), we design a difficulty-aware data construction pipeline, as shown in Fig.~\ref{fig:data-pipeline}. The pipeline leverages task instructions, screen states, and multi-source UI Agent outputs, and incorporates three complementary mechanisms: rule-based verification, structured perturbation, and intention-centric grounding correction. These components systematically generate positive samples and multiple types of negative samples. The resulting dataset provides DS-RM with fine-grained, semantically coherent, and diverse supervision signals.

\paragraph{Rule-Based Verification}
Given a task instruction--screen pair $(x, s)$, the MagicGUI agent~\citep{tang2025magicgui} produces a step-level action $a_{\text{pred}}$. We then apply deterministic UI execution rules to assess action correctness along three axes:
(1) type alignment, ensuring that the predicted action $a_{\text{pred}}$ matches the category of ground-truth action $a_{\text{gt}}$;
(2) spatial validity, requiring the predicted coordinates $(u_{\text{pred}}, v_{\text{pred}})$ to fall within valid UI-element regions $\mathcal{B}_{\text{valid}}$;
(3) semantic equivalence, verifying that the predicted action provides the correct input text or aligns with the intended operation.
Actions satisfying all rules are labeled as positive samples, while any violation yields a hard negative samples:
\[
a_{\text{pred}} \in 
\begin{cases}
	\mathcal{D}^{+}, & \text{if all constraints hold},\\[2pt]
	\mathcal{D}^{\text{hard}}, & \text{otherwise}.
\end{cases}
\]
Hard negatives represent visually plausible but operationally invalid actions, supplying DS-RM with high-precision supervision for learning strict domain-grounded interaction logic.

\paragraph{Structured Perturbation}
To systematically synthesize a broader range of negative samples, we introduce structured perturbations to successful execution paths. Two mechanisms are employed:  
(1) Instruction substitution, which replaces the original instruction with a semantically related but operationally incompatible variant $x'=\operatorname{sub}(x)$, simulating task misinterpretation;  
(2) Trajectory stitching, which concatenates segments from different tasks $\tau_i$ into a mismatched execution sequence $\tau'=\tau_1 \oplus \tau_2$, modeling contextual inconsistency and goal confusion.  

Under these perturbed inputs, MagicGUI produces actions $a_{\text{pert}}$ that may appear locally reasonable yet deviate from the intended task semantics:
\[
a_{\text{pert}} \in \mathcal{D}^{\text{easy}}.
\]
These easy negatives capture controllable failure modes such as semantic drift, partial intent errors, and cross-task mismatch, thereby enriching the diversity and coverage of the reward modeling dataset.

\paragraph{Intention-Centric Grounding Correction}
To introduce additional semantically aligned supervision while controlling for grounding noise, we incorporate actions produced by open-source UI Agents $\{\pi^{(k)}_{\text{OS}}\}$:
\[
a_{\text{OS}}^{(k)} = \pi^{(k)}_{\text{OS}}(x, s).
\]
We determine whether an action's high-level intention matches the ground truth. If the intention is correct but the grounding is inaccurate, spatial coordinates are repaired using valid UI regions and the corrected action $a_{\text{corr}}^{(k)}$ is treated as a positive sample.  
If the intention is incorrect, the sample becomes an moderate negative:
\[
a_{\text{OS}}^{(k)} \in \mathcal{D}^{\text{mid}}.
\]
This mechanism provides explicit signals for distinguishing correct and incorrect intentions, enabling DS-RM to model task-level semantic consistency.

\begin{table}[t]
	
	\centering
	\caption{Summary of agent datasets and reward datasets used for training and evaluation.}
	\begin{tabular}{ccccccc}
		\toprule
		\textbf{Agent Dataset} & \textbf{Size} & \textbf{Manual Annotation} & \textbf{Average Steps} & \textbf{Traj. Num} & \textbf{Task Instruct} \\
		\midrule
		Android Control  & 88k & \checkmark & 5.5 & 15283 & High\&Low \\
		MagicGUI-Agent-39k & 39k & \checkmark  & 4.0 & 9890 & High \\
		\midrule
		\textbf{Reward Dataset} & \textbf{Size} & \textbf{Manual Annotation} & \textbf{Data Source} & \textbf{Positive Num} & \textbf{Negative Num} & \\
		\midrule
		MagicGUI-RMS-72k & 72k & $\times$ & MagicGUI-Agent-39k & 38909 & 34013 \\
		\bottomrule
	\end{tabular}
	\label{tab:datasets}
\end{table}

\paragraph{Reward Model Dataset}
Integrating the above mechanisms yields the final reward modeling dataset:
\[
\mathcal{D}_{\text{RMS}} =
\mathcal{D}^{+}
\cup \mathcal{D}^{\text{easy}}
\cup \mathcal{D}^{\text{mid}}
\cup \mathcal{D}^{\text{hard}}.
\]
Positive samples reflect correct intentions and precise operational behavior;  
easy negatives capture explicit intention errors;  
moderate negatives represent instruction-level inconsistencies and cross-task mismatches;  
and hard negatives reflect visually plausible yet rule-violating operations.  
This construction yields the MagicGUI-RMS-72k dataset, which provides a structured and difficulty-aware sample distribution that enables DS-RM to master domain-specific execution constraints with high fidelity and provides GP-RM with well-typed inputs for global arbitration. The detailed data statistics are summarized in Table.~\ref{tab:datasets}.

\subsection{Reward-Guided Data Reflux and Co-Evolution}
\label{sec:method-reflux}

MagicGUI-RMS establishes a reward-guided data reflux mechanism that continuously improves both the UI Agent and the Domain-Specific Reward Model (DS-RM). As illustrated in Fig.~\ref{fig:magicgui-rms framework}, this mechanism operates at the step level: given a user instruction, a screen state, and the historical trajectory, the UI Agent produces an action proposal that is iteratively evaluated and corrected by DS-RM and GP-RM. The resulting signals form a closed-loop system in which the agent’s action policy and the reward models co-evolve through sustained supervision.

At each step, the UI Agent generates a candidate action
\[
a_{\text{pred}} = \pi_{\text{Agent}}(x, s, h_{1:t-1}),
\]
which is processed by DS-RM using the composite input
\[
z_{\text{DS}} = (x, s, a_{\text{pred}}, h_{1:t-1}).
\]
DS-RM outputs a binary decision $y_{\text{DS}}\in\{0,1\}$ and a domain-grounded rationale $r_{\text{DS}}$, and when $y_{\text{DS}}=0$, a corrected action $a_{\text{corr}}$ with a corresponding correction rationale $r_{\text{corr}}$:
\[
(y_{\text{DS}}, r_{\text{DS}}, a_{\text{corr}}, r_{\text{corr}})=f_{\text{DS}}(z_{\text{DS}}).
\]
This process functions as a domain-aware action rectifier that not only assesses action correctness but also proposes executable corrections aligned with domain rules.

To incorporate broader semantic reasoning and reduce the risk of domain-model bias, GP-RM performs a secondary evaluation. GP-RM receives the augmented input
\[
z_{\text{GP}} = (x, s, a_{\text{pred}}, h_{1:t-1}, y_{\text{DS}}, r_{\text{DS}}, a_{\text{corr}}, r_{\text{corr}}),
\]
and predicts whether the DS-RM decision is reliable, whether the current task has completed, and what operational intent the agent should adopt next:
\[
(y_{\text{GP}}, e_{\text{GP}}, s_{\text{GP}})=f_{\text{GP}}(z_{\text{GP}}).
\]
GP-RM thereby acts as a global oversight model that validates DS-RM’s reasoning within a wider contextual space and identifies cases where DS-RM fails to capture the correct semantics.

The final action for reflux is selected based on GP-RM’s preference between the agent’s original action and DS-RM’s corrected action:
\[
a^{\ast} = 
\operatorname*{argmax}_{a \in \{a_{\text{pred}}, a_{\text{corr}}\}}
R_{\text{GP}}(a \mid z_{\text{GP}}),
\]
where $R_{\text{GP}}(\cdot)$ denotes GP-RM’s evaluation procedure. The selected action $a^{\ast}$ is returned to the UI Agent Training Set as a high-quality supervisory label. Over time, the agent’s policy becomes increasingly aligned with the behaviors jointly endorsed by DS-RM and GP-RM, enabling stable long-term improvement driven by its own rollouts.

Disagreements between DS-RM and GP-RM form a crucial signal for refining DS-RM. When GP-RM determines that DS-RM misjudged a correct action or produced an incorrect correction, the corresponding sample is added to the RMS Training Set as a high-priority instance. These disagreement samples allow DS-RM to iteratively adjust its decision boundaries and incorporate broader contextual logic from GP-RM, enabling it to retain its domain-specific expertise while gradually gaining more global decision-making capability.

Through repeated cycles of action prediction, hierarchical evaluation, and bidirectional reflux, MagicGUI-RMS establishes a closed-loop co-evolution process. The UI Agent improves by imitating reward-validated actions, DS-RM becomes increasingly accurate through targeted disagreement-informed training, and GP-RM focuses progressively on corner cases as lower-level models mature. This reward-guided data reflux mechanism transforms the reward system from a passive evaluator into an active engine for long-term self-evolution.

\section{Experiments}

In this section, we evaluate MagicGUI-RMS and MagicGUI-Agent across a diverse set of benchmarks designed to assess step-level reward modeling and task execution. Sec.~\ref{sec:implementation-details} details our implementation setup. Sec.~\ref{sec:evaluation benchmarks} provides an overview of the evaluation benchmarks. Sec.~\ref{sec:ablation study} presents ablation studies that analyze the contribution of each system component.

\subsection{Implementation Details}
\label{sec:implementation-details}
We implement MagicGUI-Agent and MagicGUI-RMS using Qwen3-VL-8B~\citep{yang2025qwen3} as the unified vision–language backbone. MagicGUI-Agent is fine-tuned on task-oriented interaction data to enhance its multimodal perception and interface grounding capabilities, while the DS-RM adapts the same backbone on our domain-aligned reward dataset to specialize its decision boundaries. In contrast, the GP-RM employs GPT-4o~\citep{hurst2024gpt} as a high-capacity external evaluator during the early stages of self-evolution.

\textbf{Agent Model Training.}
The MagicGUI-Agent is obtained through supervised fine-tuning on a combined training corpus comprising the MagicGUI-Agent-39k dataset, which is constructed by selecting 39k complex and high-level tasks from the MagicGUI interaction corpus~\citep{tang2025magicgui}, and the AndroidControl dataset~\citep{li2024effects}. In addition to supervised fine-tuning, the agent is further improved through reinforcement fine-tuning guided by the reward functions defined in MagicGUI~\citep{tang2025magicgui}, which enhances its ability to learn both real-world mobile task execution patterns and standardized UI control behaviors.
During self-evolution, MagicGUI-Agent interacts with real screens to produce trajectories, which are then evaluated by DS-RM and GP-RM. The final reward-validated action $a^{\ast}$ is added back into the training set as a high-quality supervisory sample.  
We perform two rounds of reward-guided improvement, resulting in a significantly stronger agent trained entirely through reward reflux.
All agent models are trained with AdamW, learning rate $1\mathrm{e}\text{-}{6}$, and batch size 8.

\textbf{Reward Model Training.}
DS-RM is trained using a multi-output supervised objective that consists of four prediction targets:
the correctness label $y_{\text{DS}}$, 
the domain-specific rationale $r_{\text{DS}}$, 
the corrected action $a_{\text{corr}}$ (when applicable), 
and its accompanying explanation $r_{\text{corr}}$. 
We first fine-tune DS-RM on the MagicGUI-RMS-72k, 
which contains positive, easy-negative, moderate-negative, and hard-negative reward samples 
generated by the hierarchical construction pipeline described in Sec.~\ref{sec:method-datacon}.
Then, DS-RM undergoes a reinforcement fine-tuning stage. 
The reward function extends the grounding rewards used in MagicGUI-Agent 
and additionally incorporates DS-RM's binary classification outcomes. 
We define the correctness reward as:

\[
R_{\text{DS}} =
\begin{cases}
	+1, & {y}_{\text{DS}} = y_{\text{GT}}, \\
	-0.5, & y_{\text{GT}} = 0 \ \text{and} \ {y}_{\text{DS}} = 1, \\
	-0.2, & y_{\text{GT}} = 1 \ \text{and} \ {y}_{\text{DS}} = 0. \\
\end{cases}
\]

This reward formulation penalizes false positives more heavily, as incorrectly validating an erroneous label may introduce harmful supervision signals into the data-reflux pipeline. False negatives receive a lighter penalty, and correct predictions are rewarded with $+1$, collectively encouraging DS-RM to maintain stable and reliable decision boundaries.
During self-evolution, we conduct two rounds of disagreement-based improvement, in which DS-RM is further optimized on GP-RM verified disagreement samples collected during the self-evolution process (Sec.~\ref{sec:method-reflux}). All DS-RM experiments adopt the AdamW optimizer with a learning rate of $1\mathrm{e}\text{-}{6}$ and a batch size of~16.

GP-RM does not undergo gradient-based training. Instead, it serves as a high-capacity semantic evaluator, producing meta-reward signals $y_{\text{GP}}$, task completion predictions $e_{\text{GP}}$, 
and action-preference summaries $s_{\text{GP}}$. 
These signals are used to refine DS-RM and to supervise the UI Agent during reward-guided data reflux.

\begin{table*}[t]
	\centering
	\caption{Step-level action prediction performance on the AndroidControl dataset (AC-Low and AC-High) and the MagicGUI-Agent-39k dataset in terms of Type Match (TM) and Exact Match (EM). Bold indicates the best results.}
	\renewcommand{\arraystretch}{1.15}
	\setlength{\tabcolsep}{6pt}
	
	\begin{tabular}{
			>{\centering\arraybackslash}p{5.5cm}
			>{\centering\arraybackslash}p{1.3cm}
			>{\centering\arraybackslash}p{1.3cm}
			>{\centering\arraybackslash}p{1.3cm}
			>{\centering\arraybackslash}p{1.3cm}
			>{\centering\arraybackslash}p{1.3cm}
			>{\centering\arraybackslash}p{1.3cm}
		}
		\toprule
		\multirow{2}{*}{\textbf{Models}} &
		\multicolumn{2}{c}{\textbf{AC-Low}} &
		\multicolumn{2}{c}{\textbf{AC-High}} &
		\multicolumn{2}{c}{\textbf{MagicGUI-Agent-39k}} \\
		\cmidrule(lr){2-7}
		& TM & EM & TM & EM & TM & EM \\
		\midrule
		
		\multicolumn{7}{c}{\textit{Closed-source Models}} \\
		\midrule
		GPT-4o~\citep{hurst2024gpt}    & -   & 19.5 & -   & 20.8 & 59.6   & 14.4 \\
		Gemini 2.0~\citep{team2023gemini} & -   & 28.5 & -   & 60.2 & 57.6   &  20.7 \\
		
		\midrule
		\multicolumn{7}{c}{\textit{Open-source Models}} \\
		\midrule
		Qwen2.5-VL-7B~\citep{bai2025qwen2} & 94.1 & 85.0 & 75.1 & 62.9 & 70.2 & 32.0 \\
		UI-TARS-7B~\citep{qin2025ui}   & 95.2 & 91.8 & 81.6 & 74.4 & 63.1 & 40.9 \\
		OS-Genesis-7B~\citep{sun2025genesis} & 90.7 & 74.2 & 77.6 & 59.8 & 49.3 & 27.5 \\
		OS-Atlas-7B~\citep{wu2024atlas}    & 73.0 & 67.3 & 70.4 & 56.5 & 66.0 & 42.3 \\
		Aguvis-7B~\citep{xu2024aguvis}     & 93.9 & 89.4 & 65.6 & 54.2 & 81.1 & 61.1 \\
		
		AgentCPM-GUI~\citep{zhang2025agentcpm} & 94.4 & 90.2 & 77.7 & 69.2 & 86.3 & 63.3 \\
		\bottomrule
		\rowcolor{blue!10} MagicGUI-Agent & \textbf{97.2} & \textbf{93.5} & \textbf{84.7} & \textbf{76.3} & \textbf{88.7} & \textbf{74.1} \\
		\bottomrule
	\end{tabular}
	\label{tab:agent-result}
\end{table*}

\subsection{Evaluation Benchmarks}
\label{sec:evaluation benchmarks}

\subsubsection{MagicGUI-Agent Evaluation}

To comprehensively evaluate the effectiveness of MagicGUI-Agent, we conduct static action prediction evaluation on two datasets: the AndroidControl benchmark~\citep{li2024effects} and the MagicGUI-Agent-39k dataset, which is constructed from MagicGUI~\citep{tang2025magicgui} training data and real-world annotated interaction data. AndroidControl provides a controlled, screenshot-only setting in which the model must infer the correct action solely from the visual interface, task instruction, and action history. Following established protocols, we report results on both high-level tasks requiring multi-step, goal-directed reasoning and low-level tasks offering explicit step-by-step guidance.
To complement this controlled benchmark, the MagicGUI-Agent-39k dataset is built through a joint sampling of the MagicGUI training corpus~\citep{tang2025magicgui} and annotated interaction traces gathered from real-world mobile usage scenarios. Its distribution is substantially broader than that of AndroidControl, capturing diverse real-world workflows such as navigation, filtering, multi-screen transitions, and search operations. This diversity enables the dataset to better reflect the structural complexity and semantic variability of real-world GUI interactions.
Detailed dataset statistics are summarized in Table~\ref{tab:datasets}.
For both datasets, we follow standard evaluation protocols and assess model performance using two widely adopted metrics: Type Match (TM), which measures whether the predicted action type matches the ground truth, and Exact Match (EM), which further requires all associated parameters to be correctly predicted. Based on these metrics, we report performance on high-level and low-level tasks in AndroidControl, and evaluate the model’s overall prediction capability under real-world interaction distributions using MagicGUI-Agent-39k. 

Table~\ref{tab:agent-result} reports step-level action prediction performance on AndroidControl (AC-Low/AC-High) and MagicGUI-Agent-39k, evaluated by Type Match (TM) and Exact Match (EM). MagicGUI-Agent delivers the best overall performance across all benchmarks, achieving 97.2/93.5 (TM/EM) on AC-Low and 84.7/76.3 on AC-High. These results indicate strong robustness when moving from simpler, more canonical interactions to more challenging settings characterized by higher ambiguity and longer-horizon dependencies. Compared with the strongest open-source baseline UI-TARS-7B, MagicGUI-Agent yields consistent gains, improving AC-Low by +2.0 TM / +1.7 EM and AC-High by +3.1 TM / +1.9 EM, reflecting more reliable action-type inference and higher step-level execution accuracy. On MagicGUI-Agent-39k, MagicGUI-Agent further attains 88.7/74.1 (TM/EM). The pronounced improvement on EM suggests that MagicGUI-Agent effectively reduces near-miss errors—cases where the intended action is reasonable yet fails due to subtle execution mismatches (e.g., correct action type but incorrect target, omitted prerequisites, or fine-grained UI constraints). Notably, the consistent degradation from AC-Low to AC-High underscores that exact matching remains the dominant failure mode under increased compositionality and UI ambiguity, highlighting the inherent difficulty of precise grounding and step-wise decision making in complex interactive scenarios.

\subsubsection{MagicGUI-RMS Evaluation}

\begin{figure}[tbp]
	\centering
	\includegraphics[width=0.95\textwidth, trim=3 3 3 3, clip]{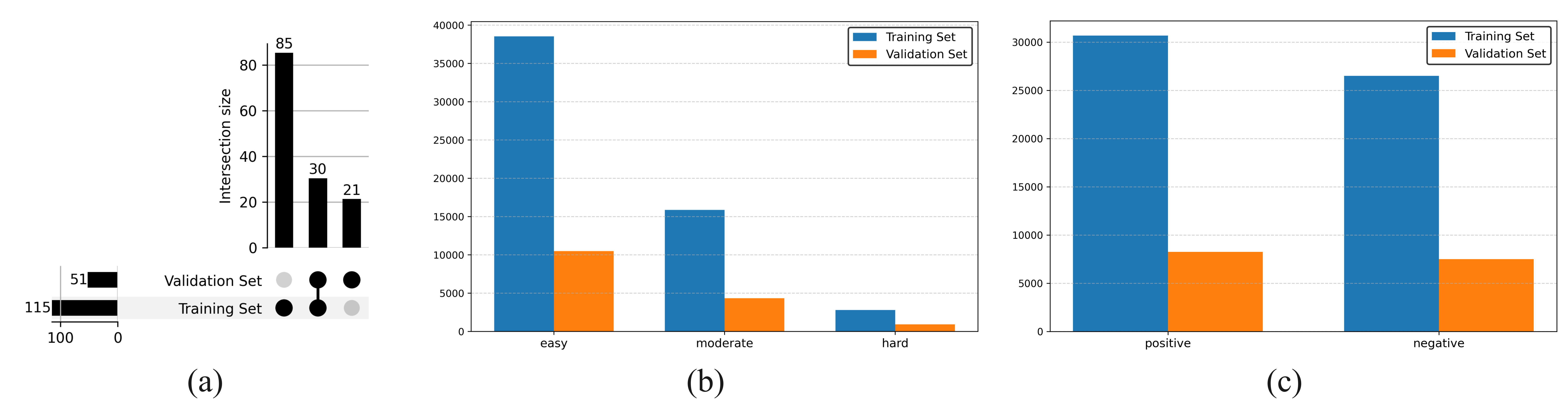}
	\caption{Statistics of the MagicGUI-RMS-72k reward dataset. (a) Distribution of application categories. (b) Distribution of difficulty levels. (c) Distribution of positive and negative samples.}
	\label{fig:rms-statistics}
\end{figure}

To evaluate MagicGUI-RMS, we conduct offline reward-model assessment using the MagicGUI-RMS-72k dataset, which is constructed through the hierarchical, difficulty-aware data pipeline described in Sec.~\ref{sec:method-datacon}. The dataset provides a balanced distribution across application domains, difficulty levels, and sample types, including positive actions, easy negatives, moderate negatives, and hard negatives, enabling comprehensive analysis of reward-model behavior under diverse conditions. 
The statistical breakdown in Fig.~\ref{fig:rms-statistics} further characterizes the MagicGUI-RMS-72k dataset by reporting the distribution of application categories, noting that the validation split contains 21 out-of-domain apps that appear only during testing, together with the difficulty-level composition across easy, moderate, and hard cases, as well as the overall balance between positive and negative samples.
We adopt step-level discrimination accuracy as the primary evaluation metric, which measures the reward model’s ability to correctly identify whether a given action is acceptable under domain rules and task semantics.

\begin{table}[t]
	\caption{Performance comparison on the MagicGUI-RMS-72k benchmark.
		The table presents different models across ALL, IDD, and OOD subsets, with results further divided into Easy, Moderate, and Hard difficulty levels. Higher values indicate better performance for all difficulty categories. The highest values in each column are highlighted in bold.}
	\centering
	\label{tab:magicgui-rms-72k}
	\renewcommand{\arraystretch}{1.35}
	\begin{tabular}{cccccccccc}
		\toprule
		\multirow{3}{*}{\textbf{Model}} 
		& \multicolumn{9}{c}{\textbf{MagicGUI-RMS-72k}} \\
		\cmidrule(lr){2-10}
		& \multicolumn{3}{c}{ALL}
		& \multicolumn{3}{c}{IDD}
		& \multicolumn{3}{c}{OOD} \\
		\cmidrule(lr){2-10}
		& Easy & Moderate & Hard 
		& Easy & Moderate & Hard 
		& Easy & Moderate & Hard \\
		\midrule
		GPT-4o~\citep{hurst2024gpt} 
		& 87.6 & 54.6 & 33.5
		& 88.2 & 54.5 & 34.3
		& 86.5 & 54.8 & 31.7 \\
		Gemini 2.0~\citep{team2023gemini} 
		& 80.4 & 53.1 & 30.7
		& 80.8 & 53.7 & 32.4
		& 79.8 & 52.0 & 26.8 \\
		\midrule
		Qwen2.5-VL-7B~\citep{bai2025qwen2}
		& 48.8 & 46.5 & 7.6
		& 45.6 & 44.6 & 7.6
		& 53.8 & 50.4 & 7.7 \\
		Qwen2.5-VL-72B~\citep{bai2025qwen2}
		& 79.1 & 51.1 & 29.4
		& 80.3 & 51.0 & 30.1
		& 77.2 & 51.3 & 27.8 \\
		Qwen3-VL-8B~\citep{yang2025qwen3}
		& 74.3 & 58.1 & 29.5
		& 75.0 & 57.4 & 29.6
		& 73.1 & 59.5 & 29.2 \\
		Qwen3-VL-32B~\citep{yang2025qwen3}
		& 70.4 & 55.2 & 33.8
		& 70.8 & 54.8 & 34.6
		& 69.8 & 55.9 & 32.0 \\
		\midrule
		
		\rowcolor{blue!10} MagicGUI-RMS
		& \textbf{93.6} & \textbf{96.1} & \textbf{68.0}
		& \textbf{93.8} & \textbf{95.9} & \textbf{69.1}
		& \textbf{93.1} & \textbf{96.6} & \textbf{65.5} \\
		
		\bottomrule
	\end{tabular}
\end{table}

Table~\ref{tab:magicgui-rms-72k} reports step-level reward discrimination accuracy on the MagicGUI-RMS-72k benchmark under ALL, in-distribution (IDD), and out-of-distribution (OOD) splits, with results further stratified by difficulty. MagicGUI-RMS consistently achieves the best performance across all settings and difficulty levels, outperforming all baseline models in every evaluation column.
On the full test set (ALL), MagicGUI-RMS attains accuracies of 93.6 / 96.1 / 68.0 on easy, moderate, and hard cases, respectively. In contrast, both proprietary and open-source baselines exhibit pronounced performance degradation as task difficulty increases, particularly on hard samples. MagicGUI-RMS maintains substantially stronger performance in these challenging cases, surpassing GPT-4o by over 30 percentage points on hard tasks, indicating superior robustness in handling long-horizon dependencies and logically constrained GUI interactions.
Similar performance trends are observed in the IDD split, where MagicGUI-RMS achieves 93.8 / 95.9 / 69.1, demonstrating reliable reward discrimination even in complex in-domain scenarios. More importantly, on the OOD split containing previously unseen applications, MagicGUI-RMS preserves high accuracy (93.1 / 96.6 / 65.5), significantly outperforming all baselines and highlighting its strong generalization capability.
Overall, these results validate the effectiveness of the hierarchical, difficulty-aware data construction pipeline and the specialized reward modeling design in enabling robust and transferable reward judgments, particularly for challenging GUI interaction scenarios.

\subsubsection{RMS-Guided Self-Evolution}

\begin{figure*}[t]
	\centering
	
	\begin{minipage}{0.5\linewidth}
		\centering
		\captionof{table}{Step-level success rates (Step SR) of MagicGUI-Agent and DS-RM across iterative self-improvement rounds. Results are reported on the ALL, IDD (In-Domain Distribution), and OOD (Out-of-Domain) subsets.
		}
		\label{tab:evolution}
		\begin{tabular}{ccccc}
			\toprule
			\multirow{2}{*}{\textbf{Round}} & \multirow{2}{*}{\textbf{Model}} & \multicolumn{3}{c}{\textbf{Step SR}} \\
			\cmidrule(lr){3-5}
			& & ALL & IDD & OOD  \\
			\midrule
			\multirow{2}{*}{Round 0} & MagicGUI-Agent &  74.1 & 73.5 & 75.0  \\
			& DS-RM &  73.6 & 73.1 & 74.4  \\
			\midrule
			\multirow{2}{*}{Round 1} & MagicGUI-Agent &  76.6 & 76.3 & 77.1  \\
			& DS-RM & 76.5 & 76.3 & 77.0 \\
			\midrule
			\multirow{2}{*}{Round 2} & MagicGUI-Agent & 78.6 & 78.4 & 78.9 \\
			& DS-RM & 78.3 & 78.1 & 78.7  \\
			\bottomrule
		\end{tabular}
	\end{minipage}
	\hfill
	\begin{minipage}{0.45\linewidth}
		\centering
		\includegraphics[width=\linewidth]{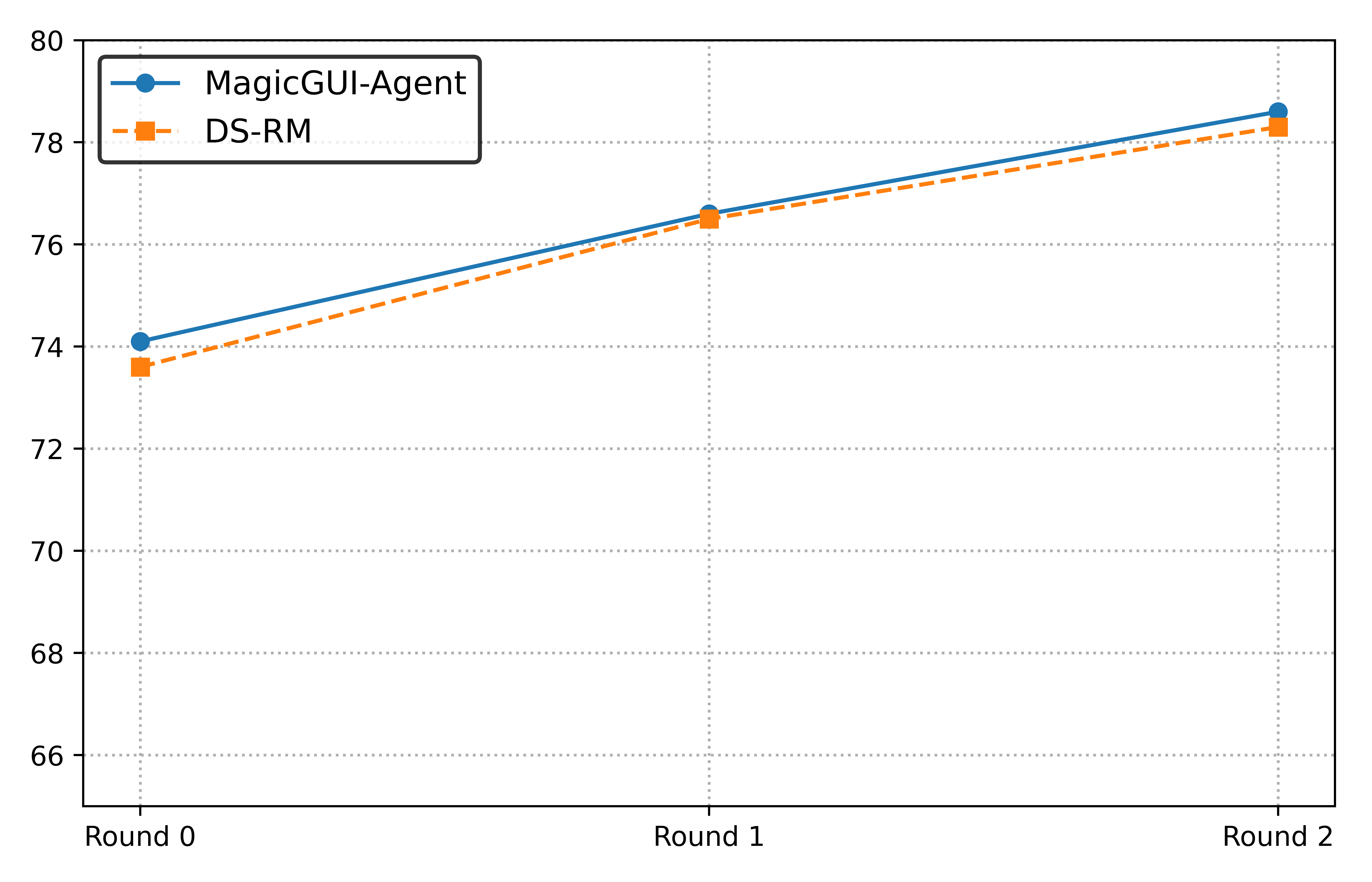}
		\caption{Performance evolution of MagicGUI-Agent and DS-RM across iterative self-improvement rounds, showing continuous gains in step-level success rate.}
		\label{fig:ui-genie-rm-evolution}
	\end{minipage}
	
\end{figure*}

To validate the effectiveness of the reward-guided reflux mechanism, we conduct multi-round self-evolution experiments built upon the MagicGUI-RMS using a collection of real-world mobile user tasks. In each iteration, MagicGUI-Agent executes 2,154 interactive episodes, whose trajectories are jointly examined by DS-RM and GP-RM. Actions that pass the reward verification are injected back into the agent’s training corpus, while GP-RM–overridden disagreement cases are added to the RMS training set to further refine the decision boundaries of DS-RM.

As shown in Table~\ref{tab:evolution} and Fig.~\ref{fig:ui-genie-rm-evolution}, the overall performance steadily improves across evolution rounds. After two RMS-guided iterations, the step-level success rate of MagicGUI-Agent increases from 74.1\% to 76.6\%, and further to 78.6\% in the second round, with consistent gains observed across ALL, IDD (In-Domain Distribution), and OOD (Out-of-Domain) subsets. DS-RM exhibits a similar trend, rising from 73.6\% to 76.5\% and ultimately reaching 78.3\%. These results demonstrate that RMS-driven self-evolution effectively mitigates model bias, enhances domain consistency, and enables sustained performance improvement across iterations.

The most pronounced improvement emerges in the first iteration, underscoring the effectiveness of reward-guided trajectory refinement in uncovering higher-quality interaction patterns in real-world tasks. As the system progressively internalizes more complex operational structures and jointly optimizes both the agent and the reward models, later iterations continue to deliver consistent performance gains. Collectively, these results demonstrate that the RMS-Guided Self-Evolution mechanism not only alleviates early-stage performance bottlenecks but also facilitates sustained capability growth through the coordinated advancement of the agent and the reward modeling components.

\subsection{Ablation Study}
\label{sec:ablation study}

\subsubsection{Effect of Reward Model Components}

\begin{table}[t]
	\caption{Component-wise ablation results of reward modeling modules on the MagicGUI-RMS-72k benchmark. The table compares the baseline vision--language model (Qwen3-VL-8B), the Domain-Specific Reward Model (DS-RM), the General-Purpose Reward Model (GP-RM), and their combination across ALL, IDD (In-Domain Distribution), and OOD (Out-of-Domain) splits. Results are reported for Easy, Moderate, and Hard difficulty levels using Step-level Success Rate (\%). Higher values indicate better performance, and the best result in each column is shown in bold.}
	
	\centering
	\label{tab:ablation-rm}
	\renewcommand{\arraystretch}{1.35}
	\begin{tabular}{cccccccccc}
		\toprule
		\multirow{3}{*}{\textbf{Model}} 
		& \multicolumn{9}{c}{\textbf{MagicGUI-RMS-72k}} \\
		\cmidrule(lr){2-10}
		& \multicolumn{3}{c}{ALL}
		& \multicolumn{3}{c}{IDD}
		& \multicolumn{3}{c}{OOD} \\
		\cmidrule(lr){2-10}
		& Easy & Moderate & Hard 
		& Easy & Moderate & Hard 
		& Easy & Moderate & Hard \\
		\midrule
		Baseline
		& 74.3 & 58.1 & 29.5
		& 75.0 & 57.4 & 29.6
		& 73.1 & 59.5 & 29.2 \\
		\midrule
		DS-RM
		& 93.1 & \textbf{97.3} & 66.7
		& 93.5 & \textbf{96.8} & 68.3
		& 92.4 & \textbf{98.2} & 63.0 \\
		GP-RM 
		& 87.6 & 54.6 & 33.5
		& 88.2 & 54.5 & 34.3
		& 86.5 & 54.8 & 31.7 \\
		\midrule
		DS-RM + GP-RM
		& \textbf{93.6} & {96.1} & \textbf{68.0}
		& \textbf{93.8} & {95.9} & \textbf{69.1}
		& \textbf{93.1} & {96.6} & \textbf{65.5} \\
		
		\bottomrule
	\end{tabular}
\end{table}

\begin{figure}[tbp]
	\centering
	\includegraphics[width=0.75\textwidth]{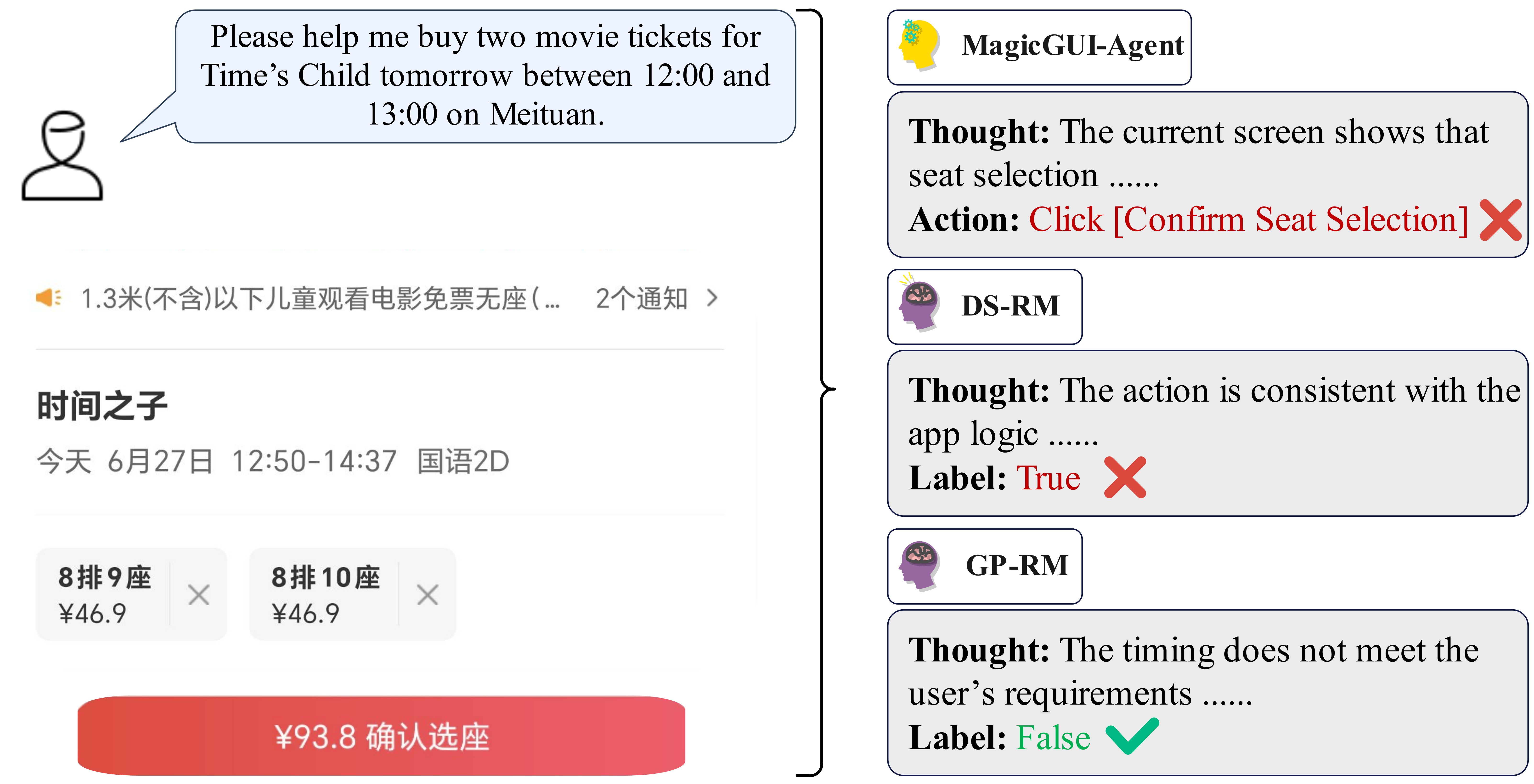}
	\caption{A representative failure case under the DS-RM–only setting. While DS-RM judges the "click [Confirm Seat Selection]" to be acceptable based on immediate interface cues, GP-RM correctly identifies the semantic mismatch between the current screening date and the user’s instruction.}
	\label{fig:gp-rm}
\end{figure}

We conduct a component-wise ablation study to quantify the contribution of each reward modeling component in MagicGUI-RMS, with results reported in Table~\ref{tab:ablation-rm}. The baseline model is Qwen3-VL-8B~\citep{yang2025qwen3}, which relies solely on generic vision--language pretraining and lacks explicit reward supervision for task execution. As shown in the table, the baseline exhibits a sharp performance drop on hard samples across ALL, IDD, and OOD splits, indicating limited capability in reasoning about complex, multi-step task execution.
The Domain-Specific Reward Model (DS-RM) leads to substantial performance gains across all splits and difficulty levels. In particular, on the ALL split, DS-RM improves the Step-level Success Rate on hard samples from 29.5 to 66.7 (+37.2), with consistent improvements observed on both IDD and OOD. These results indicate that DS-RM effectively captures task execution logic by modeling domain-specific operational constraints and interface-level action validity, enabling more accurate step-level evaluation.
The General-Purpose Reward Model (GP-RM) plays a complementary role by enforcing high-level semantic consistency. While GP-RM alone yields relatively limited improvements, its benefit is more evident on hard and out-of-domain samples, where global task understanding is required. When combined with DS-RM, GP-RM consistently further improves performance, particularly on hard cases (e.g., +2.5 on OOD--Hard), demonstrating that global semantic constraints help correct locally plausible but globally inconsistent decisions.
This effect is further illustrated in Fig.~\ref{fig:gp-rm}. Without GP-RM, the model may approve actions that are locally valid from an interface perspective but violate the overall task intent. By contrast, GP-RM correctly identifies such task-level semantic mismatches, reinforcing cross-step semantic coherence and improving robustness in complex and unseen environments.

\begin{table}[t]
	\caption{Effect of Explicit Operational Knowledge (EOK) on DS-RM performance over the MagicGUI-RMS-72k benchmark. The table compares the vanilla model, the EOK-enhanced model, and GPT-4o across ALL, IDD (In-Domain Distribution), and OOD (Out-of-Domain) subsets. Results are reported for Easy, Moderate, and Hard levels, with higher values indicating better performance and the best score in each column shown in bold.}
	\centering
	\label{tab:ablation-eok}
	\renewcommand{\arraystretch}{1.35}
	\begin{tabular}{cccccccccc}
		\toprule
		\multirow{3}{*}{\textbf{Model}} 
		& \multicolumn{9}{c}{\textbf{MagicGUI-RMS-72k}} \\
		\cmidrule(lr){2-10}
		& \multicolumn{3}{c}{ALL}
		& \multicolumn{3}{c}{IDD}
		& \multicolumn{3}{c}{OOD} \\
		\cmidrule(lr){2-10}
		& Easy & Moderate & Hard 
		& Easy & Moderate & Hard 
		& Easy & Moderate & Hard \\
		\midrule
		GPT-4o 
		& 87.6 & 54.6 & 33.5
		& 88.2 & 54.5 & 34.3
		& 86.5 & 54.8 & 31.7 \\
		\midrule
		DS-RM
		& 93.1 & 97.3 & 66.7
		& 93.5 & 96.8 & 68.3
		& 92.4 & 98.2 & 63.0 \\
		
		DS-RM + EOK
		& \textbf{97.7} & \textbf{99.3} & \textbf{96.1}
		& \textbf{97.5} & \textbf{99.4} & \textbf{95.9}
		& \textbf{98.0} & \textbf{99.1} & \textbf{96.5} \\
		
		\bottomrule
	\end{tabular}
\end{table}

\subsubsection{Impact of Explicit Operational Knowledge Injection}

In this section, we incorporate Explicit Operational Knowledge (EOK) into DS-RM through structured representations that encode hierarchical action dependencies and valid execution paths in mobile GUI tasks. With these operational priors, DS-RM gains the ability to reason over multi-step action coherence, detect prerequisite violations, and prevent reward misassignment in complex interaction flows. For example, in a task such as \emph{finding an EV charging station in Gaode Maps}, EOK captures the intended operational sequence of \emph{launching the application, opening the search bar, swiping the functional panel to reveal the corresponding category, and selecting the "Charging Station" entry.} This structured prior guides DS-RM to recognize prerequisite relationships that are not directly observable from a single screen.

As shown in Table~\ref{tab:ablation-eok}, injecting EOK leads to substantial and consistent gains across the MagicGUI-RMS-72k benchmark. The most notable improvement appears on moderate and hard cases, settings that require multi-step logical dependencies, where DS-RM+EOK delivers up to {30$\sim$33\%} higher accuracy in both in-domain and out-of-domain evaluations. These gains demonstrate that EOK effectively closes most structural reasoning failure modes observed in the vanilla reward model, while also enabling DS-RM to outperform GPT-4o by a large margin in the most challenging subsets.

Overall, these results show that EOK introduces a more faithful and discriminative representation of operational task logic, allowing DS-RM to avoid heuristic shortcuts and align its reward judgments more accurately with domain-specific execution correctness.

\section{Conclusion}
We presented MagicGUI-RMS, a multi-agent reward modeling framework that addresses the key limitations of existing GUI agents in adaptive evaluation, trajectory correction, and continual improvement. By integrating a domain-specific reward model with a general-purpose reward model, MagicGUI-RMS enables precise, interpretable, and transferable trajectory assessment. A structured reward-data construction pipeline, together with an automated data reflux mechanism, enables scalable generation and continual refinement of high-quality supervision signals, elevating reward modeling from a static post-hoc verifier to a proactive engine that continually drives agent evolution.

Extensive experiments across offline benchmarks and real-world out-of-domain environments demonstrate that MagicGUI-RMS substantially improves task accuracy, robustness, and convergence stability. These results highlight the effectiveness of MagicGUI-RMS for building autonomous, generalizable, and self-improving GUI agents, offering a scalable foundation for future research on adaptive interaction intelligence.

\bibliographystyle{delta_tuning}
\bibliography{references}

\end{document}